\documentclass[10pt,twocolumn,letterpaper]{article}

\usepackage{cvpr}
\usepackage{times}
\usepackage{epsfig}
\usepackage{graphicx}
\usepackage{amsmath}
\usepackage{amssymb}

\usepackage{tabularx}
\usepackage{multirow}

\newcommand{\argmin}{\mathop{\rm argmin}\limits}

\usepackage[pagebackref=true,breaklinks=true,letterpaper=true,colorlinks,bookmarks=false]{hyperref}

\cvprfinalcopy 

\def\cvprPaperID{997} 

\ifcvprfinal\fi
\begin{document}

\title{Visual Language Modeling on CNN Image Representations}

\author{
    Hiroharu Kato\\
    The University of Tokyo\\
    {\tt\small kato@mi.t.u-tokyo.ac.jp}
    \and
    Tatsuya Harada\\
    The University of Tokyo\\
    {\tt\small harada@mi.t.u-tokyo.ac.jp}
}

\maketitle

\begin{abstract}
    Measuring the naturalness of images is important to generate realistic images or to detect unnatural regions in images. Additionally, a method to measure naturalness can be complementary to Convolutional Neural Network (CNN) based features, which are known to be insensitive to the naturalness of images. However, most probabilistic image models have insufficient capability of modeling the complex and abstract naturalness that we feel because they are built directly on raw image pixels. In this work, we assume that naturalness can be measured by the predictability on high-level features during eye movement. Based on this assumption, we propose a novel method to evaluate the naturalness by building a variant of Recurrent Neural Network Language Models on pre-trained CNN representations. Our method is applied to two tasks, demonstrating that 1) using our method as a regularizer enables us to generate more understandable images from image features than existing approaches, and 2) unnaturalness maps produced by our method achieve state-of-the-art eye fixation prediction performance on two well-studied datasets.
\end{abstract}

\section{Introduction}
    Measuring naturalness of images is an important problem. By measuring naturalness, one can generate realistic images or detect unnatural regions in images.

    Convolutional Neural Networks (CNNs)~\cite{lecun1998gradient} extract and abstract features from raw image pixels hierarchically. Because representations they have learned for image classification are highly discriminative and generalized~\cite{donahue2013decaf, simonyan2014very}, they are an extremely important component in computer vision. However, they are known to be insensitive to the naturalness of images. For example, images generated through them appear to be strange and unrealistic~\cite{mordvintsev2015inceptionism}. Furthermore, they are susceptible to unnatural noise or artificial fabrication~\cite{goodfellow2014explaining, nguyen2014deep, szegedy2013intriguing}. Therefore, a method to measure naturalness can be complementary to CNN features.

   Despite the importance of measuring the naturalness of images, few alternative methods exist. Although many probabilistic image models have been proposed~\cite{bengio2013deep, hinton2006fast, kingma2013auto, larochelle2011neural, salakhutdinov2009deep}, most are applicable only for small image patches, not for large natural images. Moreover, they are generally built directly on raw image pixels. Although it is a preferred property for low-level image processing such as image denoising, they have insufficient capability of modeling complex and abstract naturalness as we feel.

    In this work, we assume that naturalness can be measured by predictability on high-level features during eye movement. For example, during moving of our eyes from left to right, one feels strangeness when viewing a scene that is not predicted by what we have seen along the way. This strangeness is presumably based on edges, shapes, or more semantic signals: not on pixel-level information.

    \begin{figure}[t]
        \begin{center}
            \includegraphics[bb=0 0 481 297,width=1.0\linewidth]{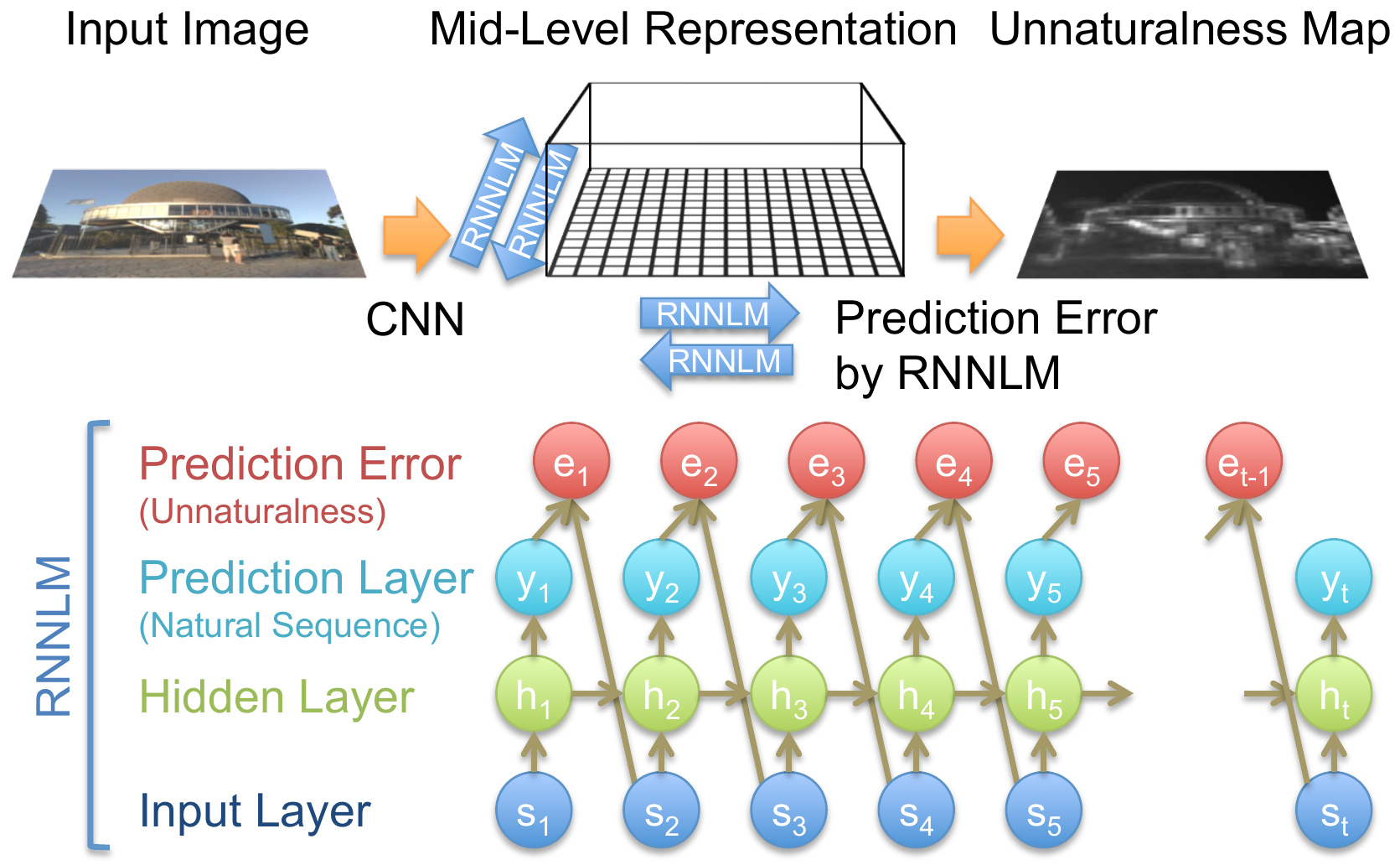}
        \end{center}
        \caption{Our pipeline to measure the unnaturalness of an image. First, mid-level representations are taken from the image using a pre-trained CNN. After they are normalized and orthogonalized, for each row and column, the naturalness of the sequence of feature vectors is evaluated using a Recurrent Neural Network Language Model (RNNLM). Naturalness by RNNLM is based on the predictability of the sequence. Finally, the ``unnaturalness map" is summed up to the unnaturalness score.}
        \label{fig:system1}
    \end{figure}

    This type of naturalness is studied extensively and is used widely in natural language processing~\cite{Rosenfeld00twodecades}. Such a method, called Language Model (LM), is applied mainly for regularizing outputs of speech recognition or machine translation systems. Given a sequence of words, LM predicts the next word from past words for each timestep and computes the naturalness of the entire sequence as the product of prediction accuracy of all timesteps. Although traditional $n$-gram models, which predict the next word from the prior $n$ words, are still prevalent, LM based on Recurrent Neural Networks (RNNs)~\cite{mikolov2010recurrent} has emerged as a favorable option because it can predict the next word using more than $n$ words.

    \begin{figure}[t]
        \begin{center}
            \raisebox{10mm}{\makebox[4mm][l]{\small{(a)}}}
            \includegraphics[width=30mm,bb=0 0 500 375]{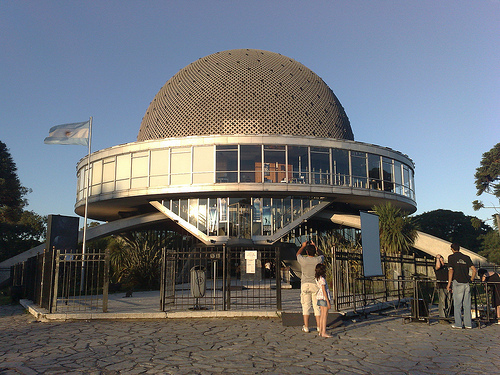} \;\;
            \raisebox{10mm}{\makebox[4mm][l]{\small{(b)}}}
            \includegraphics[width=30mm,bb=0 0 499 374]{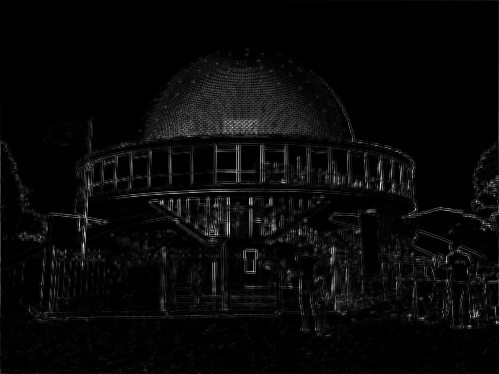} \\
            \raisebox{10mm}{\makebox[4mm][l]{\small{(c)}}}
            \includegraphics[width=30mm,bb=0 0 249 186]{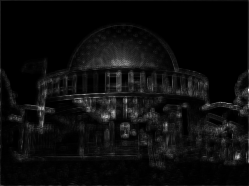} \;\;
            \raisebox{10mm}{\makebox[4mm][l]{\small{(d)}}}
            \includegraphics[width=30mm,bb=0 0 124 92]{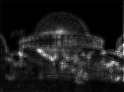} \\
            \raisebox{10mm}{\makebox[4mm][l]{\small{(e)}}}
            \includegraphics[width=30mm,bb=0 0 61 45]{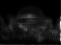} \;\;
            \raisebox{10mm}{\makebox[4mm][l]{\small{(f)}}}
            \includegraphics[width=30mm,bb=0 0 30 22]{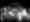} 
        \end{center}
        \caption{Unnaturalness maps computed from representations of VGGNet~\cite{simonyan2014very}: (a) Original image. (b--f) Computed from the outputs after {\tt conv1\_1}, {\tt conv2\_1,} {\tt conv3\_1}, {\tt conv4\_1}, and {\tt conv5\_1} layer. }
        \label{fig:saliency_examples}
    \end{figure}

    Motivated by the description above, we propose a method to evaluate image naturalness by building a variant of RNNLM on mid-level image representations extracted from a pre-trained CNN. RNNLM cannot accommodate two-dimensional representations. Therefore, we apply it vertically and horizontally. We designate this method as Visual Language Model on CNN (CNN-VLM). Figure~\ref{fig:system1} presents an illustration of our method. Figure~\ref{fig:saliency_examples} presents an example of unnaturalness (or prediction error) maps obtained using our method. CNN-VLM measures the naturalness using contextual information related to high-level discriminative features of CNN, which is difficult to accomplish using probabilistic image modeling on raw pixels. 

    The concept to treat image representations like words is related to the well-known Bag-of-Visual-Words~\cite{csurka2004visual, sivic2003video}. However, in our case, the representations are not quantized.
    
    We apply our method to 1) image reconstruction from image features and 2) eye-fixation prediction. The naturalness of images plays an important role in these two tasks. We briefly describe them below.

    Image reconstruction from image features is a task to visualize an image feature by reconstructing an image from it. To reconstruct interpretable images, regularizers of some kind must be imposed on images~\cite{mahendran2014understanding, simonyan2013deep}. We demonstrate empirically that using CNN-VLM as a regularizer enables us to generate more understandable images than those produced using two existing methods.

    Human eye fixation points on images were shown to be predictable using Shannon's ``self-information"~\cite{bruce2005saliency}. In fact, many attention models are explainable from the perspective of information theory~\cite{borji2013state}. Because self-information can be interpreted as unlikelihood or unnaturalness, our unnaturalness map is useful as a saliency map. We demonstrate that CNN-VLM achieves state-of-the-art performances on eye fixation prediction task in the experiment section.

    Our contributions are summarized as follows. 1) Based on the assumption that naturalness can be measured by the predictability on high-level features during eye movement, we proposed a novel method to evaluate naturalness by building a variant of RNN Language Models on pre-trained CNN representations. 2) We confirmed that using our method as a regularizer enables us to generate more understandable images from features than existing approaches. 3) We showed that unnaturalness maps produced using our method achieve state-of-the-art eye fixation prediction performance on two well-studied datasets.

\section{Related work}
    This section presents a brief review of existing approaches which are related to modeling naturalness of images. Additionally, we describe image reconstruction from features and eye fixation prediction.

    \subsection{Image modeling}
        Many probabilistic image models have been proposed~\cite{bengio2013deep, goodfellow2014generative, hinton2006fast, kingma2013auto, larochelle2011neural, salakhutdinov2009deep, goodfellow2014generative}. However, most are only applicable to small and fixed-size image patches of simple contents. To overcome this matter, Gregor \etal~\cite{gregor2015draw} used attention mechanisms and succeeded in generating very realistic images. Denton \etal~\cite{denton2015deep} applied Generative Adversarial Networks~\cite{goodfellow2014generative} in a hierarchical way. 

        Theis and Bethge~\cite{theis2015generative} proposed a scalable image model using multi-dimensional LSTMs~\cite{graves2009offline} which predict pixel values of certain locations from preceding pixels. We also use RNN for predictions like theirs. However, to capture more high-level information, we train RNN on CNN representations, not on raw pixels.

        Several vision papers explicitly use LM. Wu \etal~\cite{wu2007visual} and Tirilly \etal~\cite{tirilly2008language} trained LMs on quantized local descriptors or {\it Visual Words}. Although their approach is similar to ours, they used LMs for classification, not for measuring naturalness. Ranzato \etal~\cite{ranzato2014video} trained a language model on a small region of videos which predicts the next time frame to learn spatial-temporal video representations.

    \subsection{Image reconstruction from features}
        Reconstructing an image from its representation is an important task to understand the characteristics of the representation. Many works have addressed this problem for hand-crafted representations~\cite{d2012beyond, kato2014image, mahendran2014understanding, vondrick2012inverting, weinzaepfel2011reconstructing} and deep representations~\cite{dosovitskiy2015inverting, mahendran2014understanding, simonyan2013deep, zeiler2014visualizing}.

        Mahendran and Vedaldi~\cite{mahendran2014understanding} showed that an image can be reconstructed by gradient descent if the representation is extracted through differentiable functions. They also demonstrated that a ``natural image prior" is necessary to reconstruct interpretable images. They regularized reconstructed images to eliminate spikes in raw pixels and to be within the natural RGB range. Simonyan \etal~\cite{simonyan2013deep} adopted a similar approach and used $L_2$ regularization on images.

        Dosovitskiy and Brox~\cite{dosovitskiy2015inverting} inverted CNN features by directly learning a CNN, which translates features to images. They demonstrated that colors and rough compositions of the original image can be reconstructed.

        Our reconstruction method is based on the work by Mahendran and Vedaldi~\cite{mahendran2014understanding}. Instead of using a hand-crafted natural image prior, we use RNNLM trained on natural images as a regularizer.

    \subsection{Eye fixation prediction}
        Modeling visual attention is fundamentally important to efficiently process massive real-world data. Especially, a task to predict eye fixation points of humans has been examined extensively~\cite{borji2013state}.

        Bruce and Tsotsos~\cite{bruce2005saliency} demonstrated that eye fixation points can be predicted using Shannon's ``self-information". This information-theoretic view has been adopted for many research efforts~\cite{borji2013state}. Our method also uses a kind of self-information.

        Many recent methods are based on supervised training on an eye fixation dataset~\cite{jiang2015salicon, judd2009learning, kruthiventi2015deepfix, kummerer2014deep, liu2015predicting, vig2014large}. Ours is also a trainable flexible model. However, because it is trained in an unsupervised manner, it requires images of the target domain but does not require eye fixation data. Because making a dataset is a troublesome task, it is a favorable property for practical applications.

\section{Visual language modeling}
    As illustrated in Figure~\ref{fig:system1}, we measure the naturalness of an image using RNNLM and CNN. The pipeline of our method and corresponding sections are explained below.
    \begin{enumerate}
        \setlength{\parskip}{0cm}
        \setlength{\itemsep}{0cm}
        \item Extract a mid-level image representations from an input image using a pre-trained CNN. (Section~\ref{sec:method_CNN}.)
        \item Normalize and orthogonalize them. (Section~\ref{sec:method_preprocessing}.)
        \item Run RNNLM forward and backward along the x-axis and y-axis to obtain prediction maps and prediction error maps. (Section~\ref{sec:method_RNN}, \ref{sec:method_RNNLM}.)
        \item Sum them up and output it as the unnaturalness score. (Section~\ref{sec:method_RNNLM}.)
    \end{enumerate}

    RNNLM must be trained on natural images in advance. The training procedure is described in Section~\ref{sec:method_training}.

    We apply our method to image reconstruction from features and eye fixation prediction. We describe the details of two applications in Section~\ref{sec:method_reconstruction} and Section~\ref{sec:method_saliency}.

    \subsection{CNN and RNN}
        First, we briefly introduce CNN and RNN. We also describe their configuration in this work.

        \subsubsection{CNN}
            \label{sec:method_CNN}

            \begin{table}[t]
                \centering
                \begin{tabular}{|c||c|c|}
                    \hline
                    Model & Layer name & Output size \\
                    \hline
                    \multirow{6}{*}{AlexNet~\cite{krizhevsky2012imagenet}} & input & $227 \times 227 \times 3$ \\
                    \cline{2-3} & conv1 & $55 \times 55 \times 96$ \\
                    \cline{2-3} & conv2 & $27 \times 27 \times 256$ \\
                    \cline{2-3} & conv3 & $13 \times 13 \times 384$ \\
                    \cline{2-3} & conv4 & $13 \times 13 \times 384$ \\
                    \cline{2-3} & conv5 & $13 \times 13 \times 256$ \\
                    \hline
                    \multirow{6}{*}{VGGNet~\cite{simonyan2014very}} & input & $224 \times 224 \times 3$ \\
                    \cline{2-3} & conv1\_\{1, 2\} & $224 \times 224 \times 64$ \\
                    \cline{2-3} & conv2\_\{1, 2\}  & $112 \times 112 \times 128$ \\
                    \cline{2-3} & conv3\_\{1, 2, 3, 4\}  & $56 \times 56 \times 256$ \\
                    \cline{2-3} & conv4\_\{1, 2, 3, 4\}  & $28 \times 28 \times 512$ \\
                    \cline{2-3} & conv5\_\{1, 2, 3, 4\}  & $14 \times 14 \times 512$ \\
                    \hline
                \end{tabular}
                \caption{Layer name and output size of convolution layers. The output size is represented as ${\it height} \times {\it width} \times {\it dimension}$. ``conv1\_\{1, 2\}" signifies that there are two layers named ``conv1\_1" and ``conv1\_2".}
                \label{tab:layer_configurations}
            \end{table}

            CNN is a neural network that achieves state-of-the-art performance for image classification~\cite{krizhevsky2012imagenet, simonyan2014very}. To extract image representations, a CNN applies 2D convolution, nonlinear activation function, and downsampling in a hierarchical way. The weights of convolution kernels are learned from data to minimize classification error.

            Mid-level representations of CNNs trained on a large-scale generic image classification dataset are shown to work as a high-performance generic image feature~\cite{donahue2013decaf, simonyan2014very}. Therefore they have become the de facto standard image feature in recent years. 

            We use the outputs immediately after the convolution layers of AlexNet~\cite{krizhevsky2012imagenet} and VGGNet~\cite{simonyan2014very} to extract mid-level representations. Table \ref{tab:layer_configurations} shows layer name in the Caffe pre-trained model~\cite{jia2014caffe} and the output size of their convolution layers.

        \subsubsection{RNN}
            \label{sec:method_RNN}
            RNN is a neural network used to predict a sequence given a sequence. For each timestep $t$, hidden unit $h_t$ receives information from input $x_t$ and previous hidden unit $h_{t-1}$. Then $h_t$ passes information to output $y_t$. Because $h_t$ and $h_{t-1}$ are connected, $y_t$ depends on $x_1, x_2, ..., x_t$. Actually, RNN can make predictions using the context of infinite length.

            The most basic type of recurrent layer is formalized as follows.
            \begin{eqnarray}
                h_t = \tanh \left( W_x x_t + W_h h_{t-1} + b \right).
            \end{eqnarray}
            However, it cannot learn long-term dependencies in fact because gradients vanish in the process of flowing through many hidden-to-hidden connections. To overcome this problem, a variant of recurrent layer called Long-Short Term Memory (LSTM) was proposed~\cite{hochreiter1997long}. We used a LSTM recurrent layer defined as shown below.
            \begin{eqnarray}
                i_t &=& \sigma \left( W_{xi} x_t + W_{hi} h_{t-1} + b_i \right). \\
                f_t &=& \sigma \left( W_{xf} x_t + W_{hf} h_{t-1} + b_f \right). \\
                o_t &=& \sigma \left( W_{xo} x_t + W_{ho} h_{t-1} + b_o \right). \\
                \tilde{c}_t &=& \tanh \left( W_{xc} x_t + W_{hc} h_{t-1} + b_c \right). \\
                c_t &=& i_t * \tilde{c}_t + f_t * c_{t-1}.  \\
                h_t &=& o_t * \tanh \left( c_t \right).
            \end{eqnarray}
            Therein, $\sigma$ represents a sigmoid activation function.

            We use RNN using LSTM for sequential prediction, as described later. Concretely, we stack two LSTM layers and one linear layer to predict sequences. We set the dimensions of two LSTM layers as half of the input layer.

    \subsection{Preprocessing}
        \label{sec:method_preprocessing}
        For example, the representation after {\tt conv1} layer of AlexNet comprises $55 \times 55$ vectors of $96$ dimensions. We normalize each dimension of such vectors to have zero-mean and unit-variance. After normalization, we apply Principal Component Analysis (PCA) on it to orthogonalize and reduce dimensions by half. Parameters of normalization and PCA are learned from the training set of ILSVRC 2012 image classification dataset~\cite{ILSVRC15}.

    \subsection{Combination of CNN and RNNLM}
        \label{sec:method_RNNLM}

        Language Models (LMs) are used to measure the naturalness of a sequence. Let $s_1, s_2, ..., s_T$ to be a sequence of $D$ dimensional vectors of length $T$. The LMs decompose the probability $p(s)$ into $p(s_1)p(s_2|s_1)p(s_3|s_1, s_2) ... p(s_T|s_1, s_2, ..., s_{T-1})$. An intuitive interpretation of this is that the probability is determined by how the next vector is predictable from past vectors.

        Common LMs treat $s_t$ of one-hot vector which represents a word and assume multinoulli distribution on $p(s_t|s_1, ..., s_{t-1})$. In contrast, we use dense real-value vector $s_t$ and assume Gaussian distribution on $p(s_t|s_1, ..., s_{t-1})$. Concretely, regarding the Gaussian distribution, we assume that the mean $\mu_t$ is determined from $s_1, ..., s_{t-1}$ using RNN described in Section~\ref{sec:method_RNN}. We also assume that the variance of timestep $t$ is $\frac{T}{t}$ because where $t$ is small the model does not know much ``context" of the sequence and predicted values are less reliable. Using this assumption, $p(s_t|s_1, ..., s_{t-1})$ is rewritten by $\mathcal{N} (s_t|\mu_t, \frac{T}{t})$. The negative log likelihood of $s$ can be written as follows.
        \begin{eqnarray}
            \label{eq:negative_log}
            \textstyle -\log p(s) \propto \frac{1}{T} \sum_{t=2}^{T} \frac{t}{T} ||s_t - \mu_t||_2^2.
        \end{eqnarray}
        It is the weighted sum of squared prediction error. 

        We expand this model to 2D mid-level representation of CNN by application of RNNLM forward and backward for each row and column. We apply the same RNNLM in the same axis and direction. Therefore, there are four RNNLMs per layer. Let $f_{y,x,l}$ to be a representation immediately after layer $l$ of size $H_l \times W_l \times D_l$. Then, we define the ``unnaturalness map" $u_{y, x, l} (1 \leq y \leq H_l-1, 1 \leq x\leq W_l-1)$ of $f_l$ as follows.
        \begin{eqnarray}
            \textstyle u_{y, x, l} &=&  \textstyle \frac{x+1}{W_l}||f_{y, x+1, l} - \mu_{y, x+1, l}^\text{right}||_2^2 + \nonumber \\ 
            &&\textstyle \frac{W_l-x+1}{W_l}||f_{y, x, l} - \mu_{y, x, l}^\text{left}||_2^2 + \nonumber \\ 
            &&\textstyle \frac{y+1}{H_l}||f_{y+1, x, l} - \mu_{y+1, x, l}^\text{down}||_2^2 + \nonumber \\ 
            &&\textstyle \frac{H_l-y+1}{H_l}||f_{y, x, l} - \mu_{y, x, l}^\text{up}||_2^2.    
        \end{eqnarray}
        Therein, $\mu_{y, x, l}^\text{right}$ is predicted value of $f_{y, x, l}$ from $f_{y, 1, l}, ..., f_{y, x-1, l}$. This corresponds to scanning of an image from left to right predicting next time-step. $\mu_{y, x, l}^\text{left}$, $\mu_{y, x, l}^\text{down}$, and $\mu_{y, x, l}^\text{up}$ are also defined as the similar way. We define the total unnaturalness of $f_l$ as
        \begin{eqnarray}
            \textstyle u_l = \frac{1}{(H_l-1)(W_l-1)} \sum_{y=1}^{H_l-1} \sum_{x=1}^{W_l-1} u_{y,x,l}.
        \end{eqnarray}

        We introduce weighting parameter $\lambda_l$ for layer $l$. Then, the total unnaturalness of an image $i$ is
        \begin{eqnarray}            
            \textstyle u_i = \sum_{l} \lambda_l u_l.
        \end{eqnarray}

    \subsection{Training of RNNLM}
        \label{sec:method_training}

        To compute naturalness, we must train RNNLM in advance by minimizing $u_i$ of many natural images. We use Stochastic Gradient Descent (SGD) and Back-Propagation Through Time (BPTT) to train RNNLM. We use the training set of ILSVRC 2012 image classification dataset~\cite{ILSVRC15} for training.

        We train RNNLM on mid-level representations of AlexNet~\cite{krizhevsky2012imagenet} and VGGNet~\cite{simonyan2014very}. For AlexNet, learning rate and momentum of SGD is set, respectively, to $10$ and $0.9$. The minibatch size is set to $16$. For VGGNet, learning rate and momentum of SGD is set, respectively, to $20$ and $0.9$. The minibatch size is set to $1$. 

        For both networks, we reduce learning rate by the factor of $0.1$ for every $5000$ iterations. We stop learning after $20,000$ iterations.

    \subsection{Image reconstruction from features}
        \label{sec:method_reconstruction}

        Mahendran and Vedaldi~\cite{mahendran2014understanding} demonstrated that image features can be inverted to the original image by gradient descent (GD) if the feature extraction function comprises differentiable elements. Their key technique is the introduction of ``natural image prior" $\mathcal{R}(i)$ into their objective function. We denote the original image as $i$ and feature extraction function as $\phi(i)$. Then, using $\lambda_r$ as the weight of the regularizer, the reconstructed image $\hat{i}$ is
        \begin{eqnarray}
            \textstyle \hat{i} = \argmin_{\hat{i}} || \phi(i) - \phi(\hat{i}) ||_2^2 + \lambda_r \mathcal{R}(\hat{i}).
        \end{eqnarray}
        They used $\mathcal{R}(\hat{i})$, which keeps pixel values in the natural range and penalizes strong intensity change in neighboring pixels. Instead, we set $\mathcal{R}(\hat{i}) = u_{\hat{i}}$. Because $u_{\hat{i}}$ is differentiable, the objective function can be minimized by GD.

    \subsection{Eye fixation prediction}
        \label{sec:method_saliency}
        
        It has been suggested that humans look at locations where Shannon's ``self-information" is high~\cite{bruce2005saliency}. Because self-information is identical to the negative logarithm of probability, unnaturalness map $u_l$ can be regarded as a kind of information map that predicts salient locations.

        We use an unnaturalness map $u_l$ as a saliency map. Additionally, we apply Gaussian blur of a size of $\sigma$ to $u_l$ according to common practice~\cite{kummerer2014deep, liu2015predicting, zhang2013saliency}. Before blurring, we take the root of $u_l$ to prevent excessive expansion of peaky values. An example of unnaturalness map or saliency map $u_l$ is presented in Figure~\ref{fig:saliency_examples}.

\section{Experiments}
    In this section, we present evaluation of the effectiveness of our proposed method by application of it to two tasks: image reconstruction from features and eye fixation prediction.

    \subsection{Image reconstruction from features}
        Here we evaluate our image reconstruction method proposed in Section~\ref{sec:method_reconstruction}. First, we discuss how to evaluate reconstructed images. Then we present reconstructed images of ours and compare them with results of existing methods. Subsequently, we combine our method with the work by Dosovitskiy and Brox~\cite{dosovitskiy2015inverting} and present further improved results. Finally, we examine the role of each layer by imposing the regularizer on the target layer.

        In common with the preceding works~\cite{dosovitskiy2015inverting, mahendran2014understanding}, we  reconstructed images from the outputs of the last fully-connected layer of AlexNet and used the first one hundred images in the validation set of ILSVRC 2012 classification dataset~\cite{ILSVRC15}.

        Our model has the following hyper-parameters: the set of layers $l$ used for mid-level representations, the weight of unnaturalness map $\lambda_l$, the weight of the regularizer $\lambda_r$, and the learning rate and momentum of GD. In this section, unless otherwise noted, we set $l \in \{ {\tt conv1}, {\tt conv2}, {\tt conv3}, {\tt conv4}, {\tt conv5} \}$, $\lambda_\text{\tt conv{\it n}}=10^{-(n-1)}$ for $n \in \{1, 2, 3, 4, 5\}$ and $\lambda_r = 10$. The learning rate and momentum of GD are set, respectively, to $2^{21}$, $0.9$. Initial solution of GD is sampled from Gaussian distribution, the mean and standard deviation of which are learned by RGB values of natural images using the training set of ILSVRC 2012 classification dataset~\cite{ILSVRC15}.

        \subsubsection{Evaluation method}
            Quantitative evaluation of whether a reconstructed image is similar to the original image or not is not straightforward. Some earlier reports of the liteature~\cite{d2012beyond, weinzaepfel2011reconstructing} have provided no quantitative analysis. Because using a kind of image feature can produce an unfair comparison, mean squared error~\cite{dosovitskiy2015inverting, kato2014image, mahendran2014understanding} or correlation coefficient~\cite{vondrick2012inverting} between reconstructed images and original images have been used to date. Vondrick~\etal~\cite{vondrick2012inverting} evaluated reconstructed images by asking humans to classify them and reported that the correlation coefficient did not always match the judgments of humans.

            Therefore, in this work, we determine similarity of images by asking humans. We provide human subjects with the original image and corresponding reconstructed images. Then they select the image from reconstructed images which is the most similar to the original image. We asked one hundred people on CrowdFlower\footnote{\url{http://www.crowdflower.com/}}.

        \subsubsection{Results of reconstruction}

            \begin{figure*}[t]
                \begin{center}
                    \raisebox{6mm}{\makebox[33mm][l]{\small{(a) Original image}}}
                    \includegraphics[width=12mm,bb=0 0 227 227]{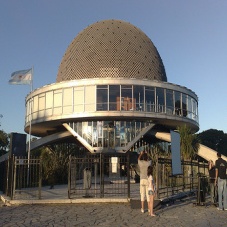}
                    \includegraphics[width=12mm,bb=0 0 227 227]{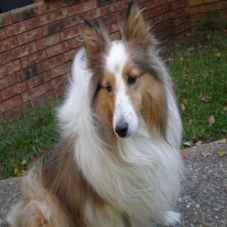}
                    \includegraphics[width=12mm,bb=0 0 227 227]{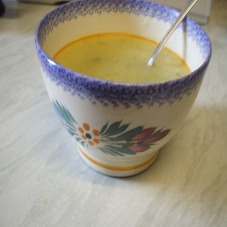}
                    \includegraphics[width=12mm,bb=0 0 227 227]{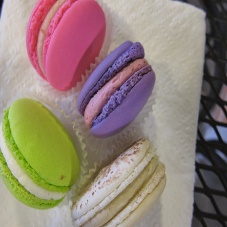}
                    \includegraphics[width=12mm,bb=0 0 227 227]{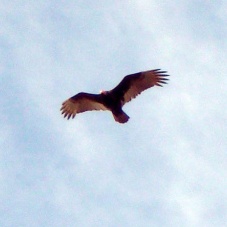}
                    \includegraphics[width=12mm,bb=0 0 227 227]{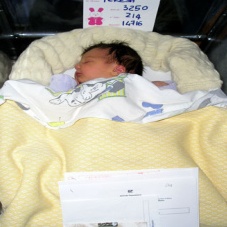}
                    \includegraphics[width=12mm,bb=0 0 227 227]{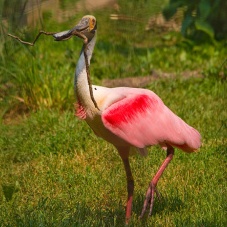}
                    \includegraphics[width=12mm,bb=0 0 227 227]{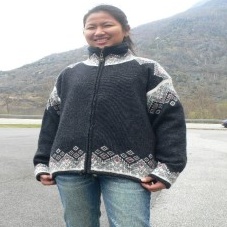}
                    \includegraphics[width=12mm,bb=0 0 227 227]{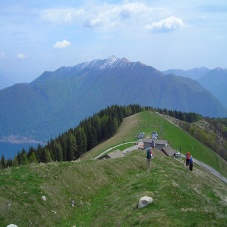}
                    \includegraphics[width=12mm,bb=0 0 227 227]{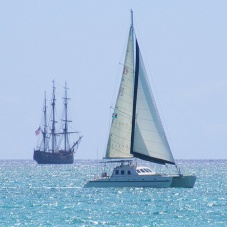}
                    \includegraphics[width=12mm,bb=0 0 227 227]{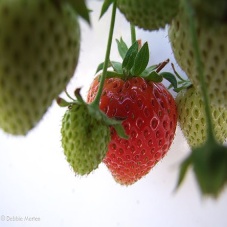}
                    
                    \raisebox{6mm}{\makebox[33mm][l]{\small{(b) Mahendran \etal~\cite{mahendran2014understanding}}}}
                    \includegraphics[width=12mm,bb=0 0 227 227]{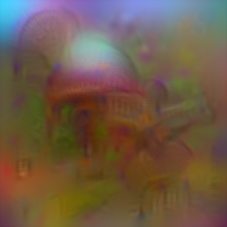}
                    \includegraphics[width=12mm,bb=0 0 227 227]{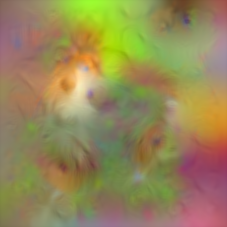}
                    \includegraphics[width=12mm,bb=0 0 227 227]{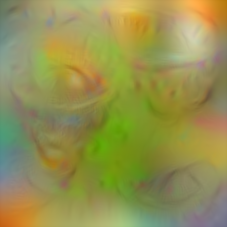}
                    \includegraphics[width=12mm,bb=0 0 227 227]{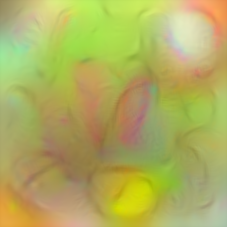}
                    \includegraphics[width=12mm,bb=0 0 227 227]{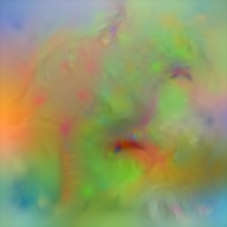}
                    \includegraphics[width=12mm,bb=0 0 227 227]{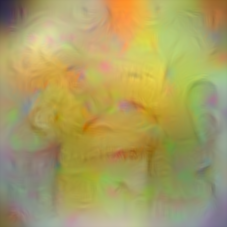}
                    \includegraphics[width=12mm,bb=0 0 227 227]{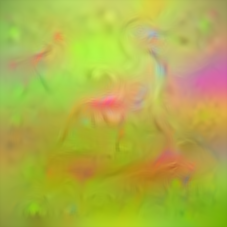}
                    \includegraphics[width=12mm,bb=0 0 227 227]{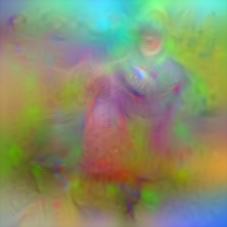}
                    \includegraphics[width=12mm,bb=0 0 227 227]{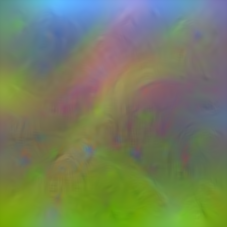}
                    \includegraphics[width=12mm,bb=0 0 227 227]{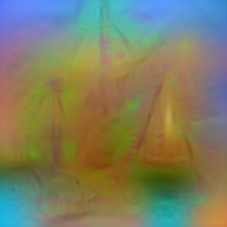}
                    \includegraphics[width=12mm,bb=0 0 227 227]{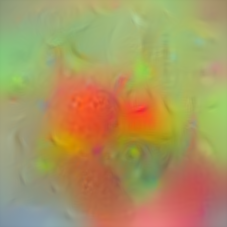}

                    \raisebox{6mm}{\makebox[33mm][l]{\small{(c) Dosovitskiy \etal~\cite{dosovitskiy2015inverting}}}}
                    \includegraphics[width=12mm,bb=0 0 128 128]{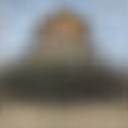}
                    \includegraphics[width=12mm,bb=0 0 128 128]{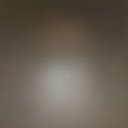}
                    \includegraphics[width=12mm,bb=0 0 128 128]{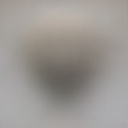}
                    \includegraphics[width=12mm,bb=0 0 128 128]{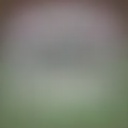}
                    \includegraphics[width=12mm,bb=0 0 128 128]{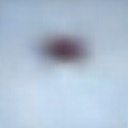}
                    \includegraphics[width=12mm,bb=0 0 128 128]{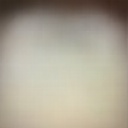}
                    \includegraphics[width=12mm,bb=0 0 128 128]{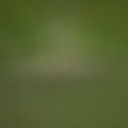}
                    \includegraphics[width=12mm,bb=0 0 128 128]{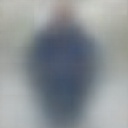}
                    \includegraphics[width=12mm,bb=0 0 128 128]{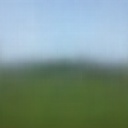}
                    \includegraphics[width=12mm,bb=0 0 128 128]{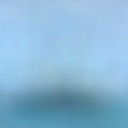}
                    \includegraphics[width=12mm,bb=0 0 128 128]{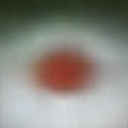}

                    \raisebox{6mm}{\makebox[33mm][l]{\small{(d) CNN-VLM (ours)}}}
                    \includegraphics[width=12mm,bb=0 0 227 227]{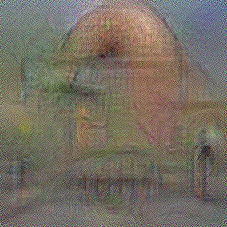}
                    \includegraphics[width=12mm,bb=0 0 227 227]{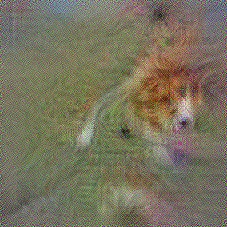}
                    \includegraphics[width=12mm,bb=0 0 227 227]{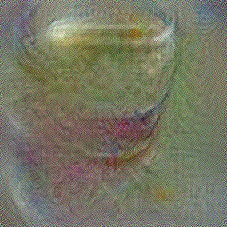}
                    \includegraphics[width=12mm,bb=0 0 227 227]{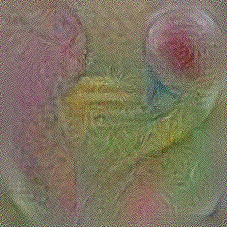}
                    \includegraphics[width=12mm,bb=0 0 227 227]{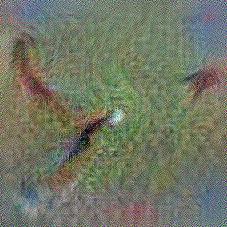}
                    \includegraphics[width=12mm,bb=0 0 227 227]{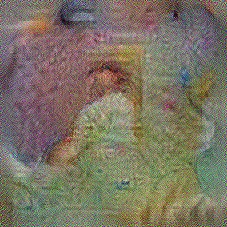}
                    \includegraphics[width=12mm,bb=0 0 227 227]{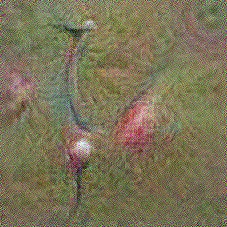}
                    \includegraphics[width=12mm,bb=0 0 227 227]{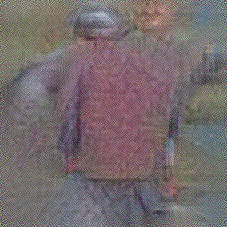}
                    \includegraphics[width=12mm,bb=0 0 227 227]{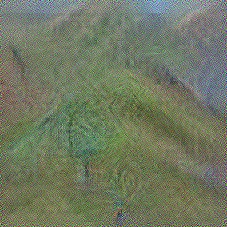}
                    \includegraphics[width=12mm,bb=0 0 227 227]{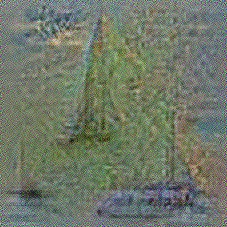}
                    \includegraphics[width=12mm,bb=0 0 227 227]{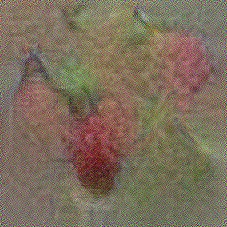}
                \end{center}
                \caption{Images reconstructed from outputs of the last fully-connected layer of AlexNet. }
                \label{fig:reconstruction}
            \end{figure*}

            \begin{table}
                \begin{center}
                    \begin{tabular}{|l||c|c|}
                        \hline
                        Method                                            & Votes           & Most similar  \\
                        \hline\hline
                        Mahendran and Vedaldi~\cite{mahendran2014understanding} & $1945$          & $4$           \\
                        Dosovitskiy and Brox~\cite{dosovitskiy2015inverting}    & $1333$          & $2$           \\
                        CNN-VLM (ours)                                          & $\mathbf{6722}$ & $\mathbf{94}$ \\
                        \hline\hline
                        Total                                                   & $10000$         & $100$         \\
                        \hline
                    \end{tabular}
                \end{center}
                \caption{Human evaluation of each reconstruction 
    method. For one hundred images, one hundred people on CrowdFlower selected the most similar image to the original image from three reconstructed images. There were a total $100 \times 100 = 10000$ votes. Actually, $94\%$ of our results are selected as the most similar ones to the original images.}
                \label{tab:experiment_quantitative}
            \end{table}

            \begin{figure}[t]
                \begin{center}
                    \includegraphics[width=12mm,bb=0 0 227 227]{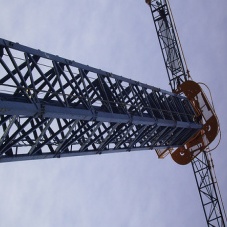}
                    \includegraphics[width=12mm,bb=0 0 227 227]{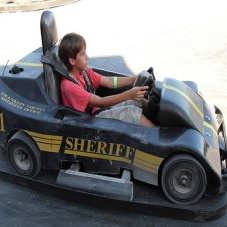}
                    \includegraphics[width=12mm,bb=0 0 227 227]{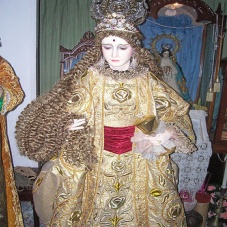}
                    \includegraphics[width=12mm,bb=0 0 227 227]{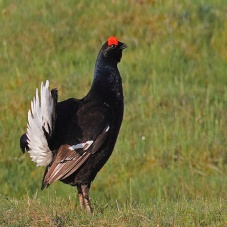} \;
                    \includegraphics[width=12mm,bb=0 0 227 227]{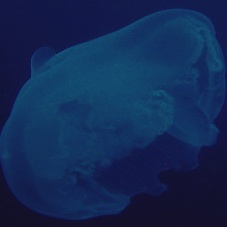}
                    \includegraphics[width=12mm,bb=0 0 227 227]{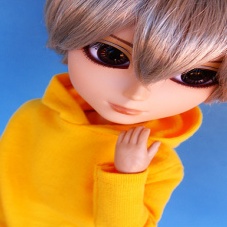}
                    
                    \includegraphics[width=12mm,bb=0 0 227 227]{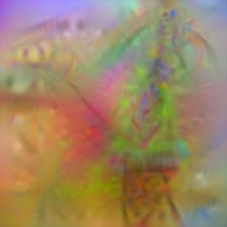}
                    \includegraphics[width=12mm,bb=0 0 227 227]{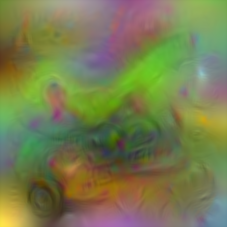}
                    \includegraphics[width=12mm,bb=0 0 227 227]{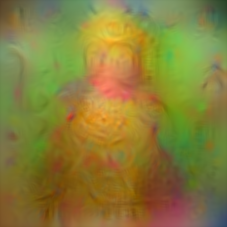}
                    \includegraphics[width=12mm,bb=0 0 227 227]{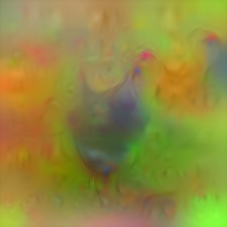} \;
                    \includegraphics[width=12mm,bb=0 0 227 227]{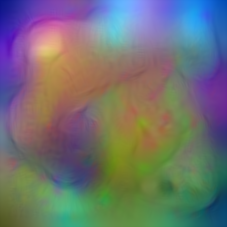}
                    \includegraphics[width=12mm,bb=0 0 227 227]{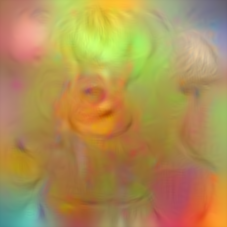}

                    \includegraphics[width=12mm,bb=0 0 128 128]{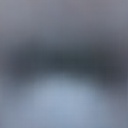}
                    \includegraphics[width=12mm,bb=0 0 128 128]{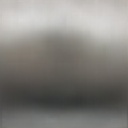}
                    \includegraphics[width=12mm,bb=0 0 128 128]{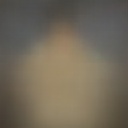}
                    \includegraphics[width=12mm,bb=0 0 128 128]{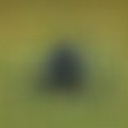} \;
                    \includegraphics[width=12mm,bb=0 0 128 128]{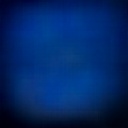}
                    \includegraphics[width=12mm,bb=0 0 128 128]{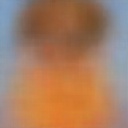}

                    \includegraphics[width=12mm,bb=0 0 227 227]{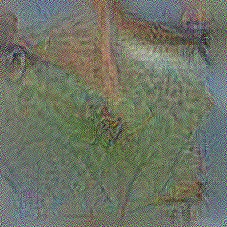}
                    \includegraphics[width=12mm,bb=0 0 227 227]{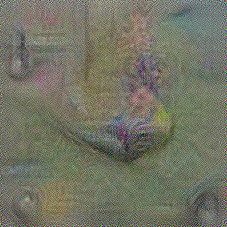}
                    \includegraphics[width=12mm,bb=0 0 227 227]{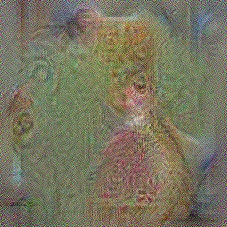}
                    \includegraphics[width=12mm,bb=0 0 227 227]{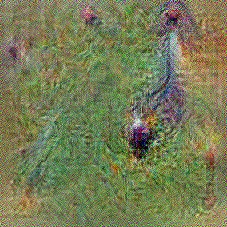} \;
                    \includegraphics[width=12mm,bb=0 0 227 227]{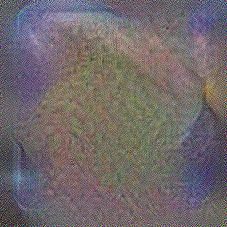}
                    \includegraphics[width=12mm,bb=0 0 227 227]{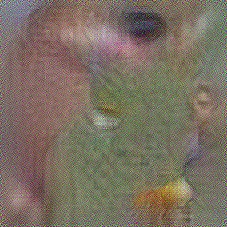}

                    \includegraphics[width=12mm,bb=0 0 227 227]{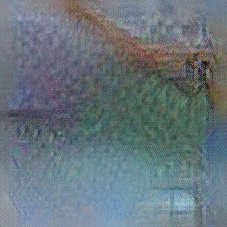}
                    \includegraphics[width=12mm,bb=0 0 227 227]{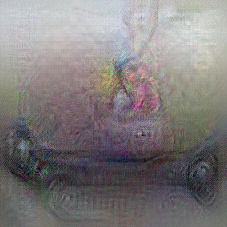}
                    \includegraphics[width=12mm,bb=0 0 227 227]{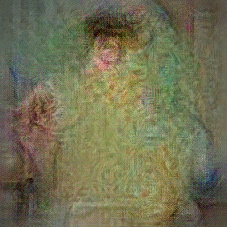}
                    \includegraphics[width=12mm,bb=0 0 227 227]{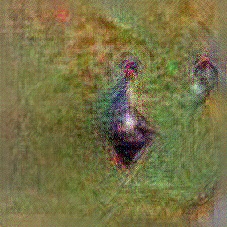} \;
                    \includegraphics[width=12mm,bb=0 0 227 227]{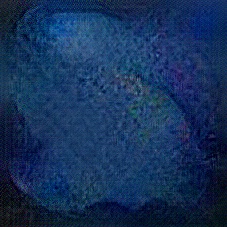}
                    \includegraphics[width=12mm,bb=0 0 227 227]{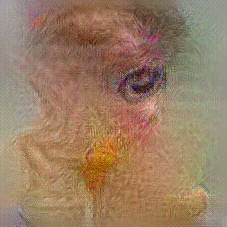}
                \end{center}
                \caption{Six cases for which our results are inferior to other methods according to human evaluation. Top row to bottom row: original image, Mahendran and Vedaldi~\cite{mahendran2014understanding}, Dosovitskiy and Brox~\cite{dosovitskiy2015inverting}, ours, and ours initialized with the results of Dosovitskiy and Brox. Left four columns: the result of Mahendran and Vedaldi is the best. Right two columns: the result of Dosovitskiy and Brox is the best. }
                \label{fig:reconstruction_best_worst}
            \end{figure}

            \begin{figure*}[t]
                \begin{center}
                    \raisebox{6mm}{\makebox[33mm][l]{\small{(e) CNN-VLM + \cite{dosovitskiy2015inverting}}}}
                    \includegraphics[width=12mm,bb=0 0 227 227]{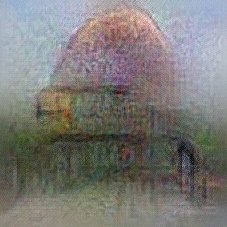}
                    \includegraphics[width=12mm,bb=0 0 227 227]{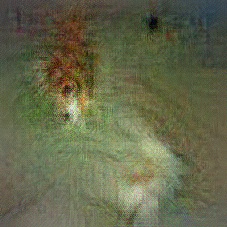}
                    \includegraphics[width=12mm,bb=0 0 227 227]{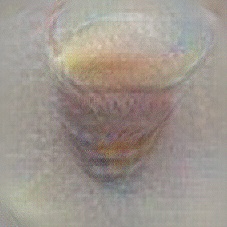}
                    \includegraphics[width=12mm,bb=0 0 227 227]{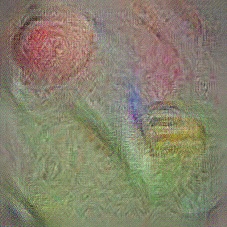}
                    \includegraphics[width=12mm,bb=0 0 227 227]{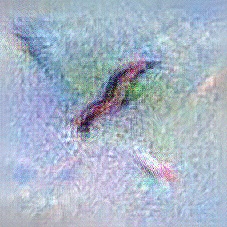}
                    \includegraphics[width=12mm,bb=0 0 227 227]{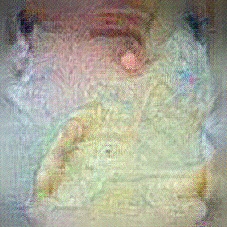}
                    \includegraphics[width=12mm,bb=0 0 227 227]{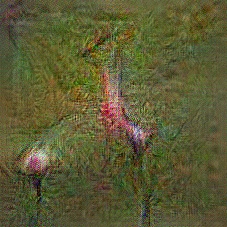}
                    \includegraphics[width=12mm,bb=0 0 227 227]{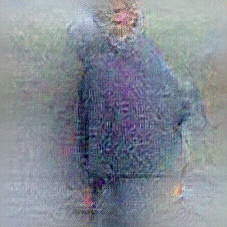}
                    \includegraphics[width=12mm,bb=0 0 227 227]{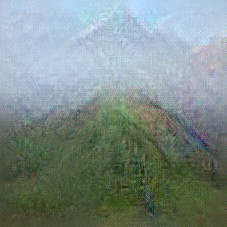}
                    \includegraphics[width=12mm,bb=0 0 227 227]{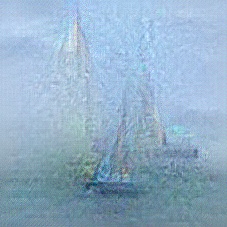}
                    \includegraphics[width=12mm,bb=0 0 227 227]{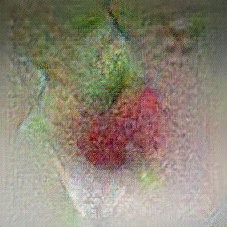}
                \end{center}
                \caption{Images reconstructed from outputs of the last fully-connected layer of AlexNet using our method. Images are initialized with the results of Dosovitskiy and Brox~\cite{dosovitskiy2015inverting}.}
                \label{fig:reconstruction_ours_dosovitskiy}
            \end{figure*}

            Figure~\ref{fig:reconstruction} depicts original images and reconstructed images by our method and two existing methods~\cite{dosovitskiy2015inverting, mahendran2014understanding}. Results of our method have clear edges rather than results of Mahendran and Vedaldi~\cite{mahendran2014understanding} because their regularizers prohibited strong change of intensity in neighboring pixels. The method presented by Dosovitskiy and Brox~\cite{dosovitskiy2015inverting} reconstructs overall shapes and colors well, although the details are lost because their method outputs an average of possible solutions. Our results appear to be the most similar to the original images. They are clear and understandable, which helps us to interpret what the image features capture.
            
            Table~\ref{tab:experiment_quantitative} presents results of quantitative evaluation by crowd sourcing. Of $100$ images, $94$ of our images are selected to the most similar images to the original images. Our results received $67.22$ \% of total votes by $100$ people, which clearly indicates the superiority of our method.
            
            Figure~\ref{fig:reconstruction_best_worst} presents six cases in which our results were judged to be inferior to two other methods. Our method is not good with reconstruction of the absolute positions and colors of objects. Additionally, our method is vulnerable to ``unnatural" images because it is trained on natural images.

        \subsubsection{Better initialization}

            The initial solution of GD is known to affect the result strongly, especially in neural networks~\cite{glorot2010understanding, sutskever2013importance}. In fact, the key to breakthroughs in deep networks resulted from smart initialization of weights~\cite{hinton2006fast}.
            
            Because results of the method by Dosovitskiy and Brox can be interpreted as the average of possible solutions, they can be good initial solutions. Figure~\ref{fig:reconstruction_ours_dosovitskiy} and the last row of Figure~\ref{fig:reconstruction_best_worst} portray reconstructed images initialized with the outputs of the method presented by Dosovitskiy and Brox.
            
            These results were improved considerably from previous results. For some images, the absolute positions and colors of objects are corrected, which indicates that the limitations of our method are mostly attributable to the initialization strategy and that they can be compensated by the method presented by Dosovitskiy and Brox.

        \subsubsection{Analysis of layers}
            \begin{figure}[t]
                \begin{center}
                    \includegraphics[width=12mm,bb=0 0 227 227]{./figure/original_227/ILSVRC2012_val_00000024.JPEG}
                    \includegraphics[width=12mm,bb=0 0 227 227]{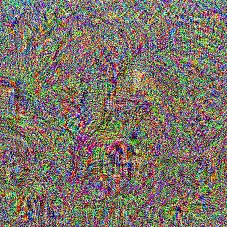}
                    \includegraphics[width=12mm,bb=0 0 227 227]{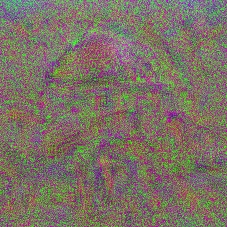}
                    \includegraphics[width=12mm,bb=0 0 227 227]{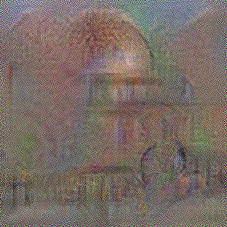}
                    \includegraphics[width=12mm,bb=0 0 227 227]{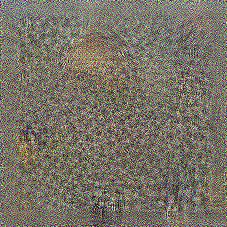}
                    \includegraphics[width=12mm,bb=0 0 227 227]{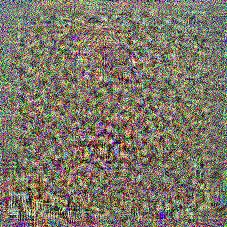} \\
                    \makebox[12mm][c]{(a)} 
                    \makebox[12mm][c]{(b)} 
                    \makebox[12mm][c]{(c)} 
                    \makebox[12mm][c]{(d)} 
                    \makebox[12mm][c]{(e)} 
                    \makebox[12mm][c]{(f)} 
                \end{center}
                \caption{Images reconstructed with regularization on various feature maps. (a) Original image. (b) No regularization. (c--f) Regularized by the naturalness of (c) the raw pixels, and the output of (d) {\tt conv1}, (e) {\tt conv2}, (f) {\tt conv3} convolution layers.}
                \label{fig:reconstruction_layer}
            \end{figure}

            Figure~\ref{fig:reconstruction_layer} shows images reconstructed using our method. In this case, the regularizer is imposed on one certain layer or raw image pixels.

            The result regularized on raw pixels, which are completely understandable, indicate the importance of modeling the naturalness of high-level features for generating realistic images, not of raw image pixels.

            The results regularized on {\tt conv2} or {\tt conv3} are less clear than the result on {\tt conv1}, which implies that the information contained by lower layers affects the naturalness of images. Higher layers can regularize more abstract information, but they are insufficient by themselves for generating images.

    \subsection{Eye fixation prediction}
        \label{sec:experiment_saliency}
        In this section, we evaluate our eye fixation prediction method proposed in Section~\ref{sec:method_saliency}. We describe details of the datasets and evaluation metrics. Subsequently, we present the results.

        Our model has two hyper-parameters: the set of layers $l$ used for mid-level representations and the kernel size of Gaussian blur $\sigma$. We use one convolution layer of VGGNet~\cite{simonyan2014very} as $l$ and set $\sigma$ to $0.030$.

        \subsubsection{Dataset}
            Many datasets are used for eye fixation prediction. Herein, we evaluate our method on two popular datasets, called MIT1003~\cite{judd2009learning} and Toronto~\cite{bruce2005saliency}. Additionally, we evaluate ours on MIT Saliency Benchmark~\cite{bylinskii2012saliency, judd2012benchmark}, which consists of two other datasets: MIT300 and CAT2000. The MIT Saliency Benchmark is an online benchmarking service. Evaluation is done in submission.

            \paragraph{MIT1003}
                MIT1003 Dataset~\cite{judd2009learning} includes 1003 images of natural indoor and outdoor scenes and corresponding eye fixation maps. It includes 779 landscape images and 228 portrait images.
            \paragraph{Toronto}
                Toronto~\cite{bruce2005saliency} includes 200 images of outdoor and indoor scenes and corresponding eye fixation maps.
            \paragraph{MIT300}
                MIT300~\cite{bylinskii2012saliency, judd2012benchmark} includes 300 images of natural indoor and outdoor scenes. Corresponding fixation maps are not provided. Because the protocol to collect this dataset is almost as the same as MIT1003, MIT1003 is useful as a training set.
            \paragraph{CAT2000}
                CAT2000~\cite{borji2015cat2000} is a recently introduced dataset. It includes $2000$ images of $20$ different image categories including action, affective, art, black \& white, cartoon, fractal, indoor, inverted, jumbled, line drawing, low-resolution, noisy, object, outdoor man-made, outdoor natural, pattern, random, satellite, sketch, and social. This dataset is challenging because most categories are not natural scenes.

        \subsubsection{Evaluation metrics}
            Eye fixation task can be interpreted as detection task to detect eye fixation point from an image. Therefore, area-under-the-curve score (AUC) of ROC curve is often used for evaluation~\cite{bruce2005saliency}. However, because humans tend to look around the center of an image, this metric assigns much value to center-biased saliency maps. To overcome this problem, shuffled AUC has been proposed~\cite{tatler2005visual, zhang2008sun}. This metric computes AUC, not on all pixels uniformly, but on center-biased eye fixation points of other images. The shuffled AUC score of centered Gaussian is about $0.5$. We use this metric for evaluation.

        \subsubsection{Results of benchmark dataset}

            \begin{figure}[t]
                \begin{center}
                    \includegraphics[bb=0 0 504 360,width=0.99\linewidth]{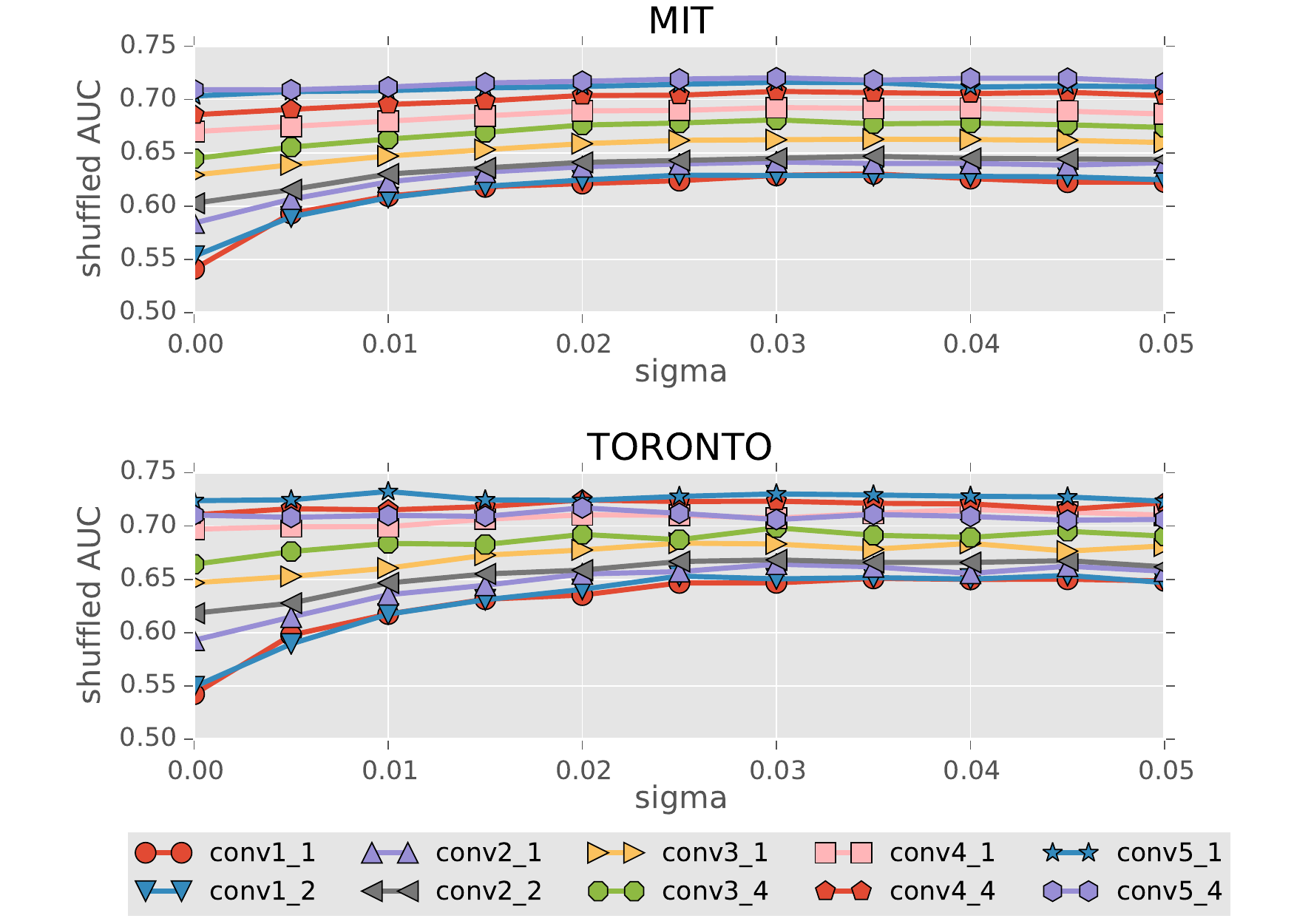}
                \end{center}
                \caption{Shuffled AUC score on MIT1003 and Toronto dataset by our method. Higher values are better. Sigma is the size of the Gaussian blur.}
                \label{fig:saliency_graph}
            \end{figure}

            \begin{table*}[t]
                \centering
                \begin{tabularx}{1\linewidth}{|>{\centering\arraybackslash}X||c c c c c c c c c c c|c|}
                    \hline
                    Dataset & Mr-CNN & AWS & BMS & CA & eDN & HFT & ICL & IS & JUDD & LG & QDCT & Ours \\
                    \hline
                    MIT1003 & .7190 & .6945 & .6939 & .6718 & .6273 & .6526 & .6667 & .6686 & .6631 & .6823 & .6686 & \textbf{.7203} \\
                    Toronto & .7236 & .7184 & .7221 & .6959 & .6292 & .6926 & .6939 & .7115 & .6901 & .6990 & .7174 & \textbf{.7323} \\
                    \hline
                \end{tabularx}
                \caption{Shuffled AUC score of each method and dataset. Scores aside from ours are cited from Liu \etal~\cite{liu2015predicting}.}
                \label{tab:mit_and_tronto}
            \end{table*}

            \begin{table*}[t]
                \centering
                \begin{tabularx}{1\linewidth}{|>{\centering\arraybackslash}X||c c c c c c c c c|c|}
                    \hline
                    Dataset & SALICON & Deep Fix & Deep Gaze I\protect \footnotemark & SalNet & Mr-CNN & AWS & CA & WMAP & IttiKoch2 & Ours \\
                    \hline
                    MIT300  & \textbf{.74} & .71 & .71 & .69 & .69 & .68 & .65 & .63 & .63 & .7096 \\
                    CAT2000 & - & .57 & - & - & - & \textbf{.62} & .60 & .60 & .59 & \textbf{.6221} \\
                    \hline
                \end{tabularx}
                \caption{Shuffled AUC score on MIT Saliency Benchmark. Scores are available online. These scores are retrieved on 30 October 2015.}
                \label{tab:mit_online}
            \end{table*}
            \footnotetext{This value is not posted online but is included in their paper.}

            Figure~\ref{fig:saliency_graph} shows the shuffled AUC score obtained using our method on MIT1003 and Toronto dataset. Various $l$ and $\sigma$ are tested. The best performing setting is $l={\tt conv5\_4}$, $\sigma=0.030$ for MIT1003 and $l={\tt conv5\_1}$, $\sigma=0.010$ for Toronto.

            Table~\ref{tab:mit_and_tronto} presents our results and existing results on the MIT1003 and Toronto datasets. We compare ours with Mr-CNN~\cite{liu2015predicting}, AWS~\cite{garcia2012relationship}, BMS~\cite{zhang2013saliency}, CA~\cite{goferman2012context}, eDN~\cite{vig2014large}, HFT~\cite{li2013visual}, ICL~\cite{hou2009dynamic}, IS~\cite{hou2012image}, JUDD~\cite{judd2009learning}, LG~\cite{borji2012exploiting} and QDCT~\cite{schauerte2012quaternion}. Ours achieved state-of-the-art score on these well-studied datasets.

            Table~\ref{tab:mit_online} presents our result on MIT300 and CAT2000. We compare ours with SALICON~\cite{jiang2015salicon}, Deep Fix~\cite{kruthiventi2015deepfix}, Deep Gaze I~\cite{kummerer2014deep}, SalNet\footnote{Unpublished work.}, Mr-CNN~\cite{liu2015predicting}, AWS~\cite{garcia2012relationship}, CA~\cite{goferman2012context}, WMAP~\cite{lopez2011scene} and IttiKoch2~\cite{itti1998model}, scores of which are available on the scoreboard of MIT Saliency Benchmark. We set $l={\tt conv5\_4}$, $\sigma=0.030$ for MIT300 and $l={\tt conv5\_1}$, $\sigma=0.030$ for CAT 2000. Our results took second place on MIT300 and first place on CAT2000.

        \subsubsection{Discussion}

            \begin{table}
                \begin{center}
                    \begin{tabular}{|c||c|c|c|}
                        \hline
                        Class & Ours & Others & Difference \\
                        \hline\hline
                        Social           & .6607 & .5764 & .0843 \\
                        Cartoon          & .6586 & .5918 & .0668 \\
                        Affective        & .6646 & .6018 & .0628 \\
                        \hline
                        Other 14 categories & -     & -     & - \\
                        \hline
                        Fractal          & .5820 & .5518 & .0302 \\
                        Low Resolution    & .5663 & .5418 & .0245 \\
                        Pattern          & .5638 & .5509 & .0129 \\
                        \hline\hline
                        Overall          & .6221 & .5755 & .0466 \\
                        \hline
                    \end{tabular}
                \end{center}
                \caption{Category-wise results on CAT2000 dataset. Others are averages of all published results available on the scoreboard. Categories for which the difference between ours and others are small or large are shown.}
                \label{tab:experiment_quantitative_saliency}
            \end{table}

            It is noteworthy that higher layers produce better results as shown in Figure~\ref{fig:saliency_graph}. Additionally, although we tried to combine all saliency maps of different layers by supervised training in our preliminary experiments, it did not produce any performance improvement. These results indicate that eye fixation points are determined exclusively from higher-level signals.

            Table~\ref{tab:experiment_quantitative_saliency} presents our score and the average of scores of submitted results available on the scoreboard of each category on the CAT2000 dataset. Compared to other methods, ours are superior for Social, Cartoon, and Affective. Presumably, that is true because images of these categories include more high-level contents such as faces or pedestrians. Ours is inferior for Pattern, Low Resolution, and Fractal because ours are trained on natural images. Images of these categories are apparently not natural. Prediction accuracy can be improved by training of RNNLM on images of these categories.

            Most top-scoring methods on MIT Saliency Benchmark are based on supervised training on eye-fixation dataset~\cite{jiang2015salicon, kruthiventi2015deepfix, kummerer2014deep, liu2015predicting}. These methods require large eye fixation dataset of target domain. If there is no dataset or the dataset is small, the performance of these models can drop. The difference of scores of DeepFix on MIT300 and CAT2000 indicates that. Ours are based on unsupervised training. Therefore it is unaffected by this problem.

\section{Conclusion}
    In this work, based on an assumption that the naturalness can be measured by the predictability on high-level features during eye movement, we proposed a novel method to measure the naturalness of an image by building a variant of RNNLMs on the CNN features. We used it as a regularizer in reconstructing images from image features. The results of experiments show that this regularizer helps to generate more feasible images than existing approaches. We found that the naturalness of lower layers is important to generate natural images. Additionally, we evaluated ``unnaturalness maps" of images as saliency maps. This was motivated by the assumption that saliency of images is based on the self-information of each location. We demonstrated that the proposed ``unnaturalness map" achieves state-of-the-art shuffled AUC scores on two well-studied eye fixation prediction datasets. It was indicated that the naturalness of higher layers predicts eye fixation points well.

    The naturalness of images, especially for large images that include rich contents, has not been studied well. Nonetheless, methods to assess and detect naturalness of images are extremely useful, as demonstrated in the experiments. We hope this work will provoke more active research in this field.

\clearpage

{
    \small
    \bibliographystyle{ieee}
    \bibliography{paper}
}

\end{document}